\documentclass[letterpaper, 10 pt, conference]{ieeeconf}  

\IEEEoverridecommandlockouts                              

\usepackage{times}
\usepackage{amssymb}
\usepackage{amsmath}
\usepackage{amsthm}
\usepackage{graphics}
\usepackage{epsfig}
\usepackage{multirow}
\usepackage{nicefrac}
\usepackage{cases}          
\usepackage{subcaption}
\usepackage{mathtools}
\usepackage{xcolor}
\usepackage{algorithm}
\usepackage{algorithmicx}
\usepackage{algpseudocode}
\usepackage{balance}
\usepackage{gensymb}

\newcommand{\mat}[0]{\begin{bmatrix}}
\newcommand{\mate}[0]{\end{bmatrix}}

\makeatletter
\let\NAT@parse\undefined
\makeatother
\usepackage{hyperref}
\hypersetup{linkbordercolor=green}
\title{\LARGE \bf
Comment on ``A Real-Time Approach for Chance-Constrained Motion Planning with Dynamic Obstacles"}
\author{Hai Zhu and Javier Alonso-Mora%
\thanks{The authors are with the Department of Cognitive Robotics, Delft University of Technology, 2628
CD, Delft, The Netherlands {\tt\small $\{$h.zhu; j.alonsomora$\}$@tudelft.nl}}%
}
\begin{document}

\maketitle              


M. Castillo-Lopez and P. Ludivig, et.al. \cite{castillo2020real} recently proposed a method for collision avoidance of uncertain dynamic obstacles. In Section VI.D, the paper compared the proposed method with several stat-of-the-art approaches using a one-horizon benchmark problem. In Fig. 2 of the paper, the comparison results were shown, as follows:
\begin{figure}[h]
    \centering
    \includegraphics[width=0.4\textwidth]{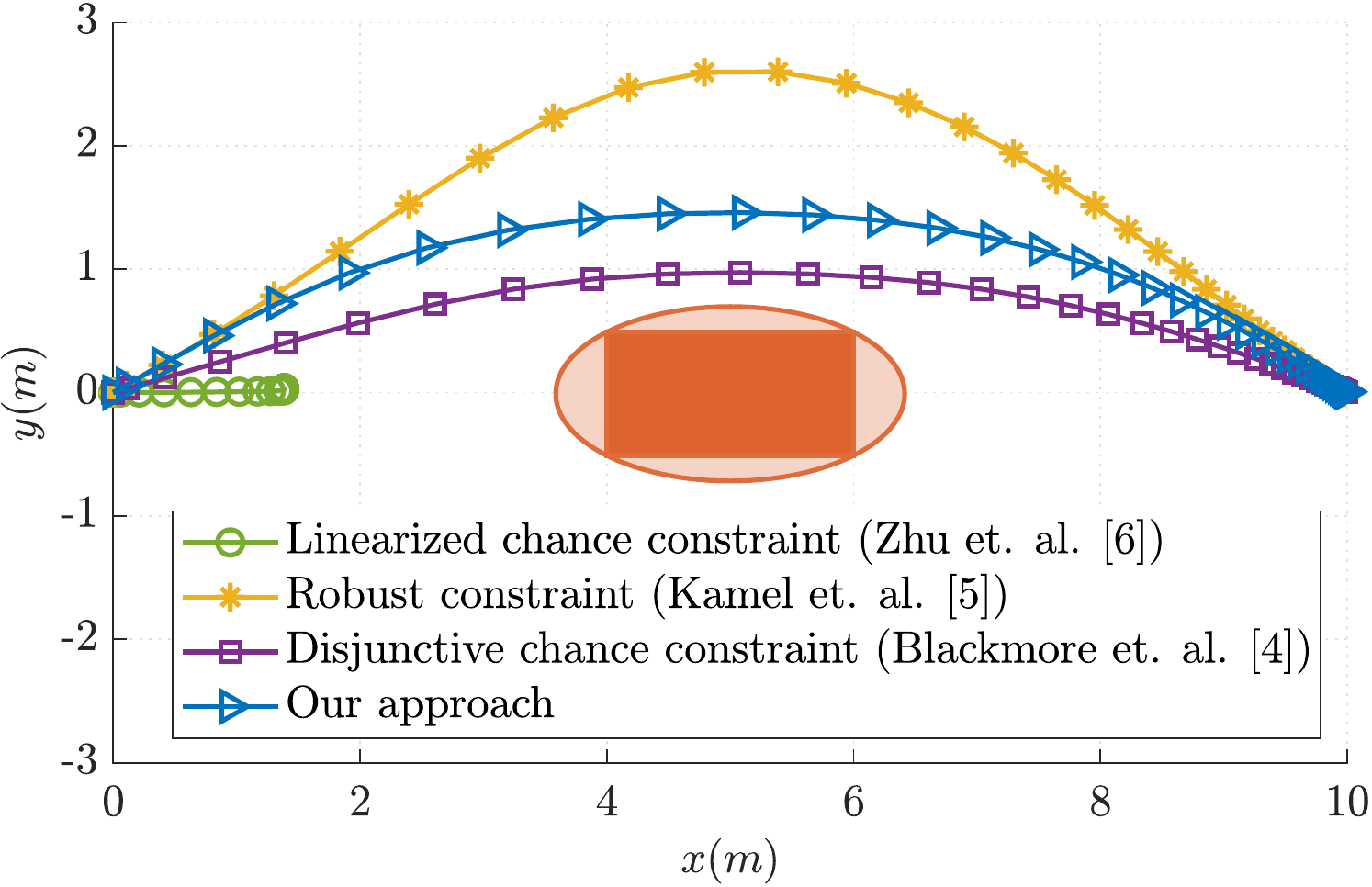}
\end{figure}

The figure shows that the method presented by Zhu et. al. in \cite{Zhu2019RAL} fell into a local minima and failed to find a feasible trajectory for the robot to its goal location. The code of the benchmark implementation can be found in \href{https://rebrand.ly/castillo_RAL2020benchmark}{\color{purple}{https://rebrand.ly/castillo\_RAL2020benchmark}}.

However, the re-implementation by \cite{castillo2020real} of the method from Zhu et. al \cite{Zhu2019RAL} is not precise. The method in \cite{Zhu2019RAL} relies on a local linearization technique to reformulate the chance constraints into deterministic constraints on the robot's and obstacle's state mean and covariance. The linearization occurs within the optimization iterations, instead of before the optimization. 

\begin{figure}[h]
    \centering
    \includegraphics[width=0.38\textwidth]{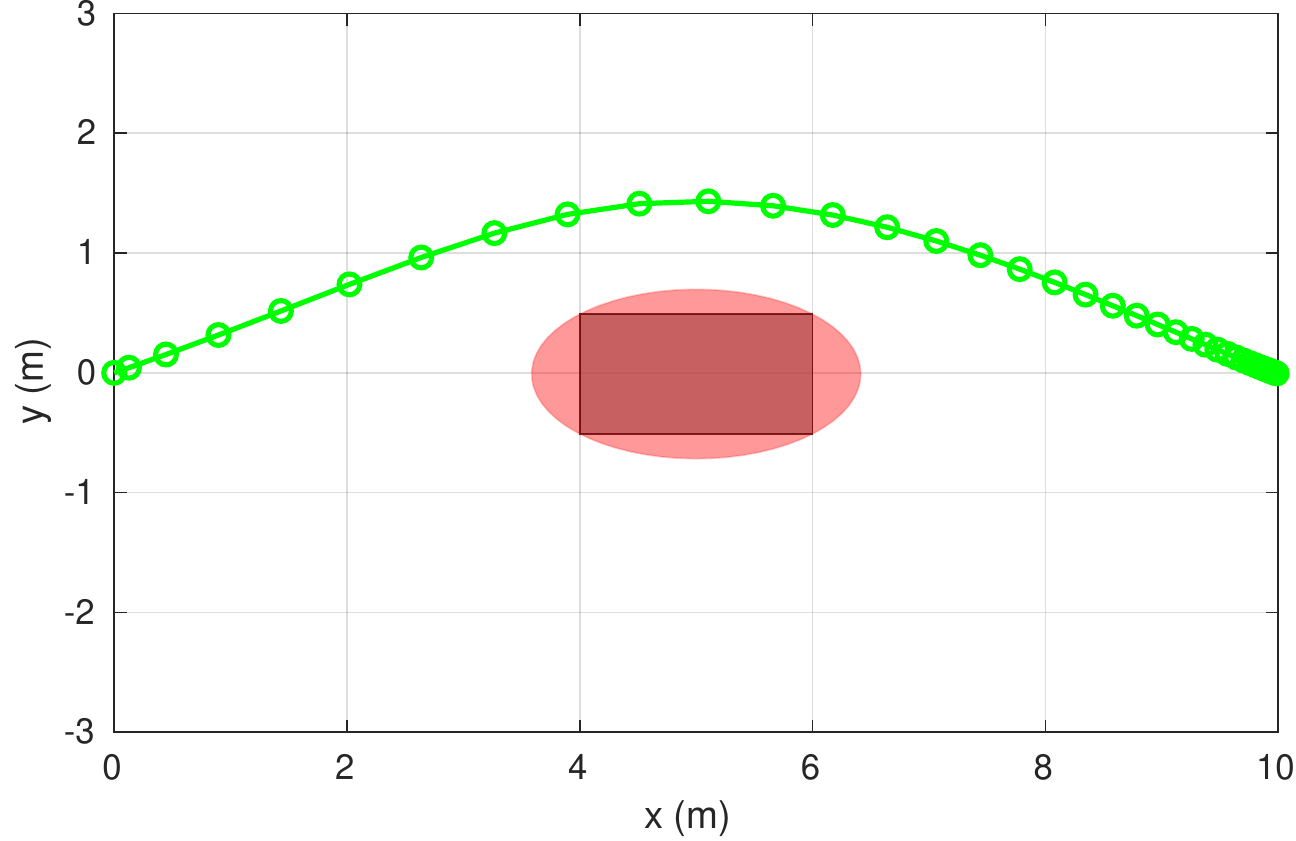}
    \caption{Results using the method of Zhu et. al. \cite{Zhu2019RAL}. }%
    \label{fig:zhu}
\end{figure}

We have reproduced the results of the benchmark problem in \cite{castillo2020real} using our method \cite{Zhu2019RAL}. Figure \ref{fig:zhu} shows the resultant trajectory of the robot.

We use the Forces Pro software \cite{FORCESNLP} to solve the non-convex optimization problem, and the computation time is 0.024 s.

The implementation of the above result can be found in \href{https://github.com/tud-amr/mrca-mav/blob/master/paper/one\_horizon\_collision\_avoidance.m}{\color{blue}{https://github.com/tud-amr/mrca-mav/blob/master/paper/one\_horizon\_collision\_avoidance.m}}.

The code of our paper \cite{Zhu2019RAL} is now available in Github: \href{https://github.com/tud-amr/mrca-mav.git}{\color{blue}{https://github.com/tud-amr/mrca-mav.git}}.

\bibliographystyle{IEEEtran}
\bibliography{ref}

\end{document}